%% file: Formatting-Instructions-LaTeX-2026.tex
\documentclass[letterpaper]{article} 
\usepackage{aaai2026}  
\usepackage{times}  
\usepackage{helvet}  
\usepackage{courier}  
\usepackage[hyphens]{url}  
\usepackage{graphicx} 
\usepackage{natbib}  
\usepackage{caption} 
\usepackage{algorithm}
\usepackage{algorithmic}

\usepackage{amsmath}
\usepackage{booktabs}
\usepackage{xcolor}
\usepackage{amssymb}
\usepackage{bm}
\usepackage{multirow}
\usepackage{tikz}
\usepackage{pgfplots}
\pgfplotsset{compat=newest}
\usetikzlibrary{arrows.meta, positioning, shapes, backgrounds}
\usepackage{xcolor,colortbl}
\usetikzlibrary{arrows.meta, positioning}

\usepackage{enumitem}
\usepackage{newfloat}
\usepackage{listings}
\DeclareCaptionStyle{ruled}{labelfont=normalfont,labelsep=colon,strut=off} 
\floatstyle{ruled}
\newfloat{listing}{tb}{lst}{}
\floatname{listing}{Listing}
\title{Active3D: Active High-Fidelity 3D Reconstruction via Hierarchical Uncertainty Quantification}
\author{
    Yan Li\textsuperscript{\rm 1}, Yingzhao Li\textsuperscript{\rm 2}, Gim Hee Lee\textsuperscript{\rm 1} \\
}
\affiliations{
    \textsuperscript{\rm 1}National University of Singapore\\
    \textsuperscript{\rm 2}Harbin Institute of Technology\\

}

\usepackage{bibentry}

\begin{document}

\maketitle

\begin{abstract}
In this paper, we present an active exploration framework for high-fidelity 3D reconstruction that incrementally builds a multi-level uncertainty space and selects next-best-views through an uncertainty-driven motion planner. 
We introduce a \emph{hybrid implicit–explicit representation} that fuses neural fields with Gaussian primitives to jointly capture global structural priors and locally observed details. 
Based on this hybrid state, we derive a \emph{hierarchical uncertainty volume} that quantifies both implicit global structure quality and explicit local surface confidence. 
To focus optimization on the most informative regions, we propose an \emph{uncertainty-driven keyframe selection} strategy that anchors high-entropy viewpoints as sparse attention nodes, coupled with a \emph{viewpoint-space sliding window} for uncertainty-aware local refinement. 
The planning module formulates next-best-view selection as an \emph{Expected Hybrid Information Gain} problem and incorporates a risk-sensitive path planner to ensure efficient and safe exploration. 
Extensive experiments on challenging benchmarks demonstrate that our approach consistently achieves state-of-the-art accuracy, completeness, and rendering quality, highlighting its effectiveness for real-world active reconstruction and robotic perception tasks.
\end{abstract}

\begin{links}
    \link{Website}{https://yanyan-li.github.io/project/vlx/active3d}
\end{links}


\section{Introduction}~\label{sec:intro}
Visual-based 3D reconstruction~\cite{newcombe2011kinectfusion,whelan2015elasticfusion,dai2017bundlefusion,li2020structure} aims to infer the geometry and appearance of previously unseen scenes from 2D imagery, making it a fundamental problem in both computer vision and robotics.
Depending on how the sensor moves, reconstruction methods can be clustered into two categories: passive and active. Passive systems process streams of RGB~\cite{schonberger2016structure} or RGB-D~\cite{li2022graph} frames to jointly estimate six-degree-of-freedom (6-DoF) camera motions and fuse the measurements into sparse or dense 3D models, under the assumption of a fixed, user-driven path. In contrast, active reconstruction frameworks~\cite{aloimonos1988active,chen2011active} integrate next-best-view (NBV) planning~\cite{peralta2020next} to autonomously select subsequent viewpoints to maximize information gain~\cite{isler2016information,kirsch2019batchbald} and ensure comprehensive surface coverage.
In addition to accurate geometry, next-generation intelligent robots demand dense 3D models with \emph{high fidelity} and photometric consistency for downstream tasks.

\begin{figure}[t]
\centering
\resizebox{0.49\linewidth}{!}{
\begin{tikzpicture}
\begin{axis}[
    width=0.9\linewidth,
    height=6cm,
    xlabel={PSNR},
    ylabel={C.R.(\%)},
    xmin=0, xmax=37,
    ymin=73, ymax=102,
    grid=both,
    grid style={dashed, gray!30},
    tick label style={font=\large},
    label style={font=\large},
    legend style={font=\tiny, at={(0.5,-0.25)}, anchor=north, legend columns=3},
    scatter/classes={
        ours={mark=o,draw=blue!70!black,fill=blue!30},
        activesplat={mark=diamond*,draw=orange!80!black,fill=orange!50},
        activegamber={mark=star,draw=brown!80!black,fill=brown!50},
        naruto={mark=pentagon*,draw=brown!80!black,fill=brown!40},
        loopsplat={mark=triangle*,draw=cyan!80!black,fill=cyan!50},
        monosgs={mark=triangle*,draw=blue!80!black,fill=blue!30},
        co_slam={mark=triangle*,draw=green!60!black,fill=green!40},
        nice_slam={mark=triangle*,draw=yellow!50!black,fill=yellow!50},
        splatam={mark=diamond*,draw=violet!70!black,fill=violet!50},
        tsdf={mark=square*,draw=gray!70!black,fill=gray!40},
        anm={mark=*,draw=red!60!black,fill=red!40},
        anm_s={mark=*,draw=red!70!black,fill=red!70}
    }
]
\addplot[scatter,only marks,scatter src=explicit symbolic]
coordinates {
    (36,98.1)    [ours]
    (28.3,96.3)  [activesplat]
    (27.8,95.15) [activegamber]
    (27.5,97.56) [naruto]
    (31.2,85.47) [loopsplat]
    (34.5,79.84) [monosgs]
    (23.5,91.22) [co_slam]
    (25.5,84.17) [nice_slam]
    (24.8,76.12) [splatam]
    (0,81.46)   [tsdf]
    (0,96.6)  [anm_s]
    (0,97.2)    [anm]
};
\node[font=\large, blue!60!black] at (axis cs:32,98.1) {OURS};
\node[font=\large, orange!80!black] at (axis cs:21.5, 95.3) {ActiveSplat};
\node[font=\large, brown!70!black] at (axis cs:28,93.3) {ActiveGamber};
\node[font=\large, brown!80!black] at (axis cs:22 ,97.5) {NARUTO};
\node[font=\large, cyan!70!black] at (axis cs:31.4,84.0) {LoopSplat};
\node[font=\large, blue!70!black] at (axis cs:32.5,78) {MonoGS};
\node[font=\large, green!50!black] at (axis cs:23.8,89.5) {CO-SLAM};
\node[font=\large, yellow!50!black] at (axis cs:25.8,82.4) {NICE-SLAM};
\node[font=\large, violet!70!black] at (axis cs:25,74.5) {SplaTAM};
\node[font=\large, gray!60!black] at (axis cs:5.5,80.7) {TSDF-Fusion};
\node[font=\large, red!70!black] at (axis cs:5,94) {ANM-S};
\node[font=\large, red!50!black] at (axis cs:5,95.9) {ANM};

\end{axis}
\end{tikzpicture}}
\includegraphics[width=0.45\linewidth, trim={260 40 0 0},clip]{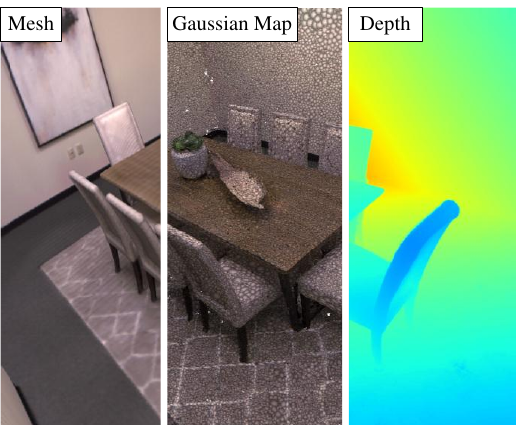}
\caption{Performance on the Replica dataset. 
Left: Comparison of rendering quality (PSNR) versus reconstruction completeness (C.R.) across state-of-the-art methods. 
Right: Qualitative outputs of our method including reconstructed mesh, Gaussian map, and estimated depth.}
\label{f2}
\end{figure}
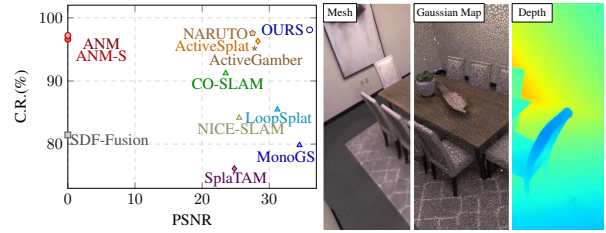

The conventional active reconstruction problem~\cite{isler2016information,huang2018active} is typically cast as an exploration task: select the sequence of viewpoints that will most effectively reveal detailed scene geometry and appearance. Early approaches leverage occupancy‐grid~\cite{elfes2013occupancy} or voxel‐based~\cite{wu2014quality} maps and frontier‐driven exploration to push the boundary between known and unknown space, ensuring that new measurements continually reduce map uncertainty~\cite{lee2022uncertainty}. 
However, since these approaches focus solely on geometric uncertainty, the resulting reconstructions are ill-suited for high-quality novel‐view rendering and often lack the photometrically consistent details required for downstream tasks. 
Recent advances in scene representation have revealed two complementary paradigms: \emph{implicit neural fields}~\cite{mildenhall2021nerf,barron2022mip} and \emph{explicit parameterizations} such as 3D Gaussian Splatting~\cite{kerbl20233dgs,li2024geogaussian}, both achieving impressive performance in novel-view synthesis and surface reconstruction. Implicit models encode continuous neural fields that excel at capturing global structure, while explicit Gaussians faithfully preserve observed geometry and fine details. However, existing active frameworks typically adopt only one of these paradigms. Implicit-based active methods~\cite{yan2023active,kuang2024active} leverage neural priors for view planning, but their continuous fields tend to hallucinate missing surfaces (e.g., transparent or mirrored areas), leading to persistent high uncertainty and planner oscillation. Conversely, GS-based active approaches~\cite{11037548,jin2025activegs} directly reflect observations into the map, providing reliable local geometry but lacking the ability to reason about occluded or unseen regions, resulting in suboptimal exploration coverage.

These complementary strengths and limitations motivate a hybrid implicit–explicit formulation for active reconstruction, unifying global priors and local textured surface within a single information-theoretic planning framework. 
First, given a posed RGB-D stream, Active3D constructs a \textbf{hybrid implicit–explicit scene state} and derives a \textbf{hierarchical uncertainty map} to jointly quantify global structural entropy and local surface uncertainty. 
Based on this hybrid uncertainty, the planner is further proposed to formulate \emph{next-best-view selection} as an Expected Hybrid Information Gain (\textbf{EHIG}) optimization and executes viewpoint-aware trajectory planning. Keyframes are promoted via a dual-uncertainty intersection criterion, selecting viewpoints that observe regions where both implicit and explicit uncertainties are high. This establishes a sparse attention mechanism over the hybrid scene state. A viewpoint-space sliding window then performs uncertainty-aware local refinement of Gaussian primitives with respect to implicit priors, maintaining global–local consistency throughout the reconstruction process. Our contributions are summarized as follows:
\begin{itemize}[leftmargin=*]
    \setlength{\itemsep}{0pt}
    \setlength{\parskip}{0pt}
    \setlength{\parsep}{0pt}
    \item We propose a \emph{hybrid implicit–explicit scene representation} for active 3D reconstruction, unifying neural fields and Gaussian primitives into a joint entropy minimization framework and introducing the \emph{Hybrid Scene State Entropy}. 

    \item We design a \emph{hierarchical uncertainty map} that fuses global implicit variance, local depth residuals, local photometric residuals, and temporal SDF changes via Bayesian fusion, providing a principled multi-scale signal to drive exploration and refinement.

    \item We formulate next-best-view planning as an \emph{Expected Hybrid Information Gain (EHIG)} problem, combining global structural exploration and local detail preservation with risk-aware path optimization. 
    \item {We introduce a viewpoint-aware keyframe selection strategy driven by the intersection of implicit and explicit uncertainties, anchoring high-information regions as sparse attention nodes in the hybrid map. Integrated with a spatial (non-temporal) sliding window, this enables uncertainty-aware local refinement and consistent reconstruction of the hybrid scene state.}
\end{itemize}


\section{Related Work}~\label{sec:related_work}

\paragraph{Neural Implicit and Explicit Representation.}
Traditionally, 3D reconstructed models have been represented using various geometric formats, including meshes~\cite{kazhdan2006poisson,li2021rgb}, surfels~\cite{whelan2015elasticfusion,stuckler2014multi}, and truncated signed distance fields (TSDF)~\cite{osher2004level,izadi2011kinectfusion}. 
With the advent of differentiable radiance fields, these representations have been significantly extended to support high-quality novel view synthesis. In particular, NeRF~\cite{mildenhall2021nerf} have emerged as a powerful paradigm for photorealistic rendering and scene understanding.
Specifically, iMAP~\cite{sucar2021imap} utilizes MLP as the only scene representation for both tracking and mapping. To address the over smoothed reconstruction problem of only-MLP representation in large-scale environments, NeuralRecon~\cite{sun2021neuralrecon} integrates neural TSDF volumes with learned features to enhance 3D reconstruction quality in indoor scenes. Similarly, ConvONet~\cite{peng2020convolutional} predicts occupancy probabilities in 3D space using 3D convolutional architectures~\cite{cciccek20163d,ronneberger2015u,niemeyer2020differentiable}, combining the strengths of spatially aware feature encoding and implicit shape modeling.

In contrast to implicit and hybrid approaches, explicit representations directly encode scene geometry and appearance in structured forms such as voxel grids~\cite{muller2022instant} or Gaussian primitives~\cite{kerbl20233dgs}, enabling efficient rendering and fast optimization. Plenoxels~\cite{fridovich2022plenoxels} replace MLPs with a sparse voxel grid that stores density and spherical harmonics coefficients. TensoRF~\cite{chen2022tensorf} further improves scalability and memory efficiency by applying low-rank tensor decomposition. More recently, 3D Gaussian Splatting~\cite{kerbl20233dgs,li2024geogaussian} introduces a point-based explicit method where each Gaussian encodes position, orientation, scale, and radiance attributes, supporting high-fidelity rendering with real-time performance and continuous surfaces.

\paragraph{Active High-quality 3D Modeling.}
Active reconstruction methods~\cite{yan2023active,kuang2024active,pan2022activenerf,11037548,jin2024gs,feng2024naruto,chen2025activegamer} autonomously select viewpoints during iterative mapping to maximize coverage and reconstruction quality. NeRF-based NBV strategies~\cite{lee2022uncertainty,pan2022activenerf} use pixel-wise rendering variance as uncertainty cues, while FisherRF~\cite{jiang2024fisherrf} introduces 
Fisher information for view planning. ANM~\cite{yan2023active} maintains weight-space uncertainty in a continually learned neural field, and 
NARUTO~\cite{feng2024naruto} extends this paradigm 
to 6-DoF exploration in large-scale scenes.

Recently, Gaussian primitives have been adopted for active scene modeling. 
ActiveGAMER~\cite{chen2025activegamer} incorporates rendering quality into the information gain metric. GS-Planner~\cite{jin2024gs} detects unobserved regions 
in the Gaussian map and employs a sampling-based NBV policy. HGS~\cite{xu2024hgs} proposes an adaptive hierarchical planning strategy balancing global and local refinement. ActiveSplat~\cite{11037548} extends Gaussian-based SLAM 
to active mapping with decoupled viewpoint orientation. 

Uncertainty estimation plays a central role in NBV selection. NeRF-based methods typically derive voxel or pixel-wise variance from density fields~\cite{pan2022activenerf,lee2022uncertainty}, while Gaussian-based methods rely on observation completeness or visibility priors~\cite{jin2024gs,11037548}. 
In contrast, we fuse \emph{global implicit variance}, \emph{local surface residuals}, and \emph{temporal SDF variation}, constructing a hierarchical uncertainty map that simultaneously guides global exploration and local refinement.

\section{Methodology}~\label{sec:method}

\begin{figure*}[!t]
    \centering
    \includegraphics[width=\linewidth]{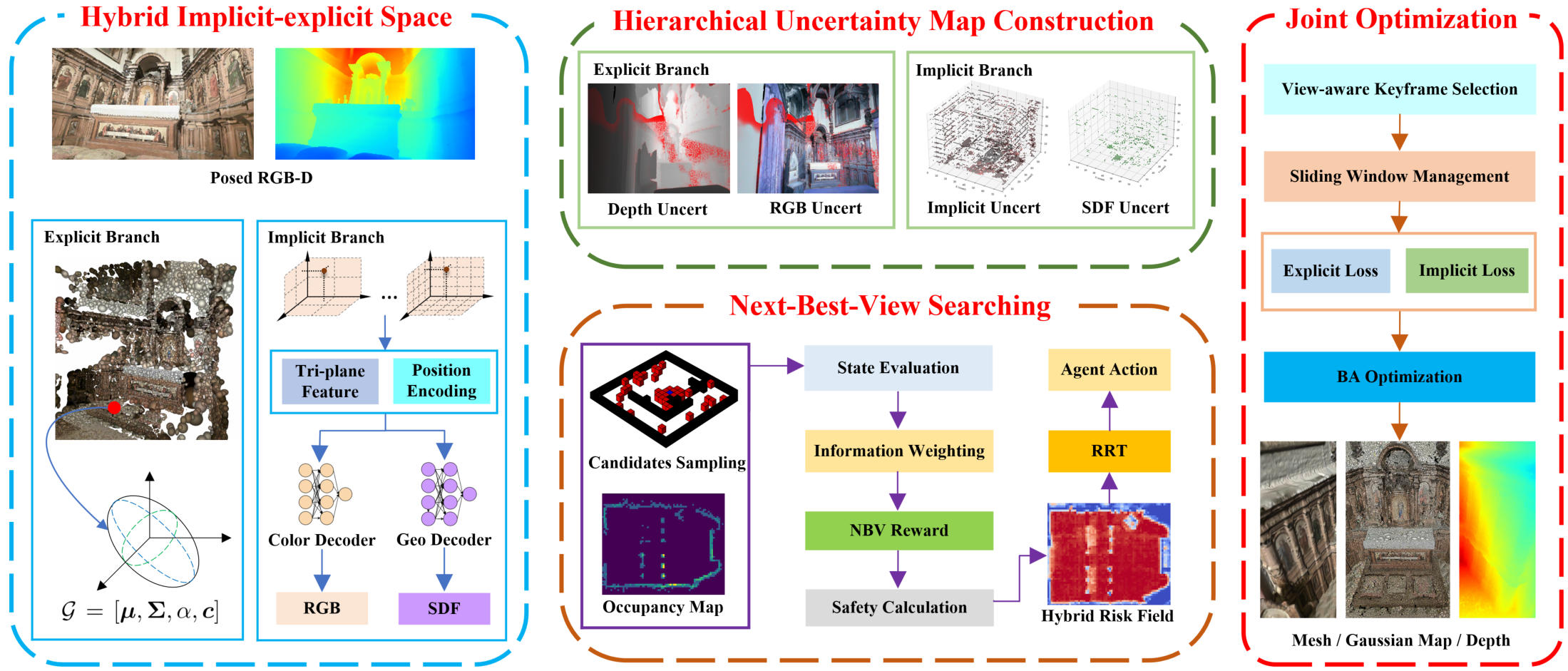}
    \caption{Our method processes the RGB-D stream through dual explicit and implicit reconstruction branches. The explicit branch projects data into a 3D Gaussian model, while the implicit branch employs an encoder-decoder architecture to regress RGB values and SDF. Subsequently, the discrepancy between the rendered RGB-D and the GT RGB-D is computed. Another mlp predicts global uncertainty, while temporal variations on the SDF surface are characterized to derive uncertainty for the hybrid explicit-implicit representation. This representation then drives NBV selection and path planning. Finally, keyframes are selected within a sliding window for joint optimization of the explicit and implicit maps.}
    \label{fig:framework}
\end{figure*}

In the active reconstruction task, the core of the problem is to decide the position and orientation of the $i^{th}$ viewpoint based on the information captured by the previous posed RGB-D stream $\mathcal{S}_{i-1} = \{ \mathbf{S}_k \}, \mathbf{S}_k=[I_k, D_k, \mathbf{T}_{c_k,w}, \mathbf{K}], k\in[0,1,\dots,i-1]$. Therefore, the problem can be defined as determining how to leverage the previously posed RGB-D stream to guide the selection of the current viewpoint in order to achieve high-quality reconstruction. This process first involves the data organization of the previous RGB-D stream, followed by quantifying the historical information to evaluate the current reconstruction state and predicting potential information gain. By modeling the scene coverage, uncertainty distribution, and geometric consistency from $\mathcal{S}_{i-1}$, the system can actively plan the next viewpoint that maximizes scene completeness and reconstruction fidelity. Fig.~\ref{fig:framework} depicts the algorithm's workflow.

\subsection{Hybrid Implicit-explicit Space}
To simultaneously capture continuous global priors and 
high-quality local surface, we construct a 
\textbf{hybrid implicit–explicit space} that integrates 
implicit neural fields with explicit  Gaussian primitives.
Given a posed RGB-D observation 
$\mathbf{S}_k = [I_k, D_k, \mathbf{T}_{c_k,w}, \mathbf{K}]$, 
this hybrid space provides a unified state representation 
for incremental active reconstruction.

\paragraph{Definition of Hybrid Scene State.}

We introduce a state formulation for the incremental active reconstruction task, where the state $\mathcal{M}_k$ at step $k$ is designed to represent the currently reconstructed portion of the scene:
\begin{equation}
\mathcal{M}_k = \{\mathcal{F}_\theta, \mathcal{G}_k\},
\label{eq:hybrid_state}
\end{equation}
where $\mathcal{F}_\theta : \mathbb{R}^3 \to \text{SDF}$ 
is the implicit neural field, and 
$\mathcal{G}_k = \{G_i\}_{i=1}^{N_k}$ is the set of 3D Gaussian primitives.
Each primitive $G_i$ is parameterized as $
G_i = (\mu_i, \Sigma_i, \alpha_i, c_i) $,
where $\mu_i \in \mathbb{R}^3$ is the mean position, 
$\Sigma_i \in \mathbb{R}^{3 \times 3}$ the covariance, 
$\alpha_i \in [0,1]$ the opacity, and $c_i \in \mathbb{R}^3$ the color vector. And for the implicit neural field, we employ a One-blob encoder~\cite{wang2023co,muller2019neural} to extract deep features from input point clouds.
The implicit representation subsequently maps world coordinates $\mathbf{x} \in \mathbb{R}^3$ to SDF values and color attributes via the MLP:
\begin{equation}
\begin{split}
    &s = f_{\tau} \big( \gamma(\mathbf{x}), \mathcal{V}_\alpha (\mathbf{x} \big))  \label{eq:sdf_mlp} \\
\end{split}
\end{equation}
where $\gamma(\mathbf{x})$ denotes tri-plane decomposition of spatial coordinates, and $\mathcal{V}_\alpha (\mathbf{x})$ represents position feature vectors obtained through volumetric trilinear interpolation. The function $f_{\tau}(\cdot)$ corresponds to the geometry decoder.

\paragraph{Hybrid State Quantification.}
At state $k$, the key objective is to quantify the \textbf{current scene knowledge} and guide the next-best-view  selection. This hybrid formulation bridges \emph{global structural exploration} driven by $\mathcal{F}_\theta$ and \emph{local high-fidelity surface} enabled by $\mathcal{G}_k$. Casting NBV planning as an \emph{expected hybrid information gain} optimization, 
we formalize active reconstruction in a probabilistic information-theoretic context.

We define the voxel-wise hybrid entropy as:
\begin{equation}
H_{\text{hybrid}}(v) = 
\lambda_{\text{imp}} H[p_{\mathcal{F}_\theta}(v)] + 
\lambda_{\text{exp}} H[p_{\mathcal{G}_k}(v)],
\label{eq:hybrid_entropy}
\end{equation}
where $H[p]$ denotes Shannon entropy and $\lambda_{\text{imp}}, \lambda_{\text{exp}}$ balance global priors and local observations.

The NBV reward for $\mathbf{c}$ is accumulated over all visible voxels:
\begin{equation}
R(\mathbf{c}) = 
\sum_{v \in \mathcal{V}_{\mathbf{c}}} 
w(v|\mathbf{c}) (1 - O(v)),
\label{eq:nbv_reward}
\end{equation}
where $\mathcal{V}_{\mathbf{c}}$ is the set of voxels visible from 
$\mathbf{c}$, and $O(v)$ is the occupancy probability used to 
discount free-space ambiguity.

\subsection{Hierarchical Uncertainty Map Construction}
\label{sec:uncertainty_map}
To drive the hybrid NBV objective in Eq.~\ref{eq:hybrid_entropy}, we construct a hierarchical uncertainty volume $\mathcal{V}_u \in \mathbb{R}^{L \times W \times H}$ that fuses \textbf{global implicit priors}, \textbf{local view-dependent surface}, and \textbf{temporal consistency cues}. Each voxel $v$ stores a scalar $u(v) \in \mathbb{R}^+$ representing the hybrid reconstruction confidence.
\paragraph{Global Structure Uncertainty.}
The implicit branch $\mathcal{F}_\theta$ encodes a continuous SDF-based representation that provides \emph{global structural entropy}. We approximate 
per-voxel variance using an uncertainty head $f_\delta(\cdot)$:
\begin{equation}
u_{\text{imp}}(v) = 
\phi \big(f_\delta(\gamma(\mathbf{x}_v), 
\mathcal{V}_\alpha(\mathbf{x}_v)) \big),
\label{eq:global_uncertainty}
\end{equation}
where $\mathbf{x}_v$ denotes the voxel center, $\gamma(\cdot)$ is the tri-plane encoder, and $\phi(\cdot)$ applies a softplus normalization. Upon receiving new observations, the structural uncertainty is updated, encouraging coverage-driven exploration and mitigating local greedy behavior during the early stages of mapping.

\paragraph{View-dependent Local Uncertainty.}
The explicit Gaussian map $\mathcal{G}_k$ provides \emph{local observation entropy} through photometric and geometric residuals. At each step, we select top-$K$ high-uncertainty candidate viewpoints $\mathcal{C}_{\text{high}}$ and compute depth and color errors:
\begin{align}
E^{\text{depth}}_t &= 
\big\| D^{\text{render}}_t - D^{\text{gt}}_t \big\| 
\odot M_t, \\
E^{\text{rgb}}_t &= 
\sum_{c \in \{R,G,B\}}
\big\| I^{\text{render}}_{t,c} - I^{\text{gt}}_{t,c} \big\| 
\odot M_t,
\end{align}
where $M_t$ masks valid pixels. 
The 2D errors are back-projected into the 3D voxel space to estimate the uncertainty of the local surface using the following formulation:
\begin{equation}
u_{\text{exp}}(v) = 
\frac{1}{|\mathcal{C}_{\text{high}}|}
\sum_{t \in \mathcal{C}_{\text{high}}}
\mathcal{P}(E^{\text{depth}}_t, 
E^{\text{rgb}}_t; v),
\label{eq:view_uncertainty}
\end{equation}
where $\mathcal{P}(\cdot)$ denotes voxel-wise backprojection with bilinear interpolation.

\paragraph{Temporal Variation Uncertainty.}
To detect emerging surfaces and inconsistencies, we evaluate SDF changes between consecutive keyframes: $ \Delta S_t = S_t - S_{t-1} $. 
According to the varying states of surfaces, define masks for new surfaces, geometry changes, and novel free space:
\begin{equation}
\left\{
\begin{aligned}
M_{\text{new}} &=\ \mathbb{I}(0{\leq}S_t{\leq}\tau_s) 
\odot \mathbb{I}(\Delta S_t{>}\tau_n), \\
M_{\text{change}} &=\ \mathbb{I}(|\Delta S_t|{>}\tau_c), \\
M_{\text{free}} &=\ \mathbb{I}(S_t{>}\tau_f) 
\odot \mathbb{I}(S_{t-1}{<}{-}\tau_f).
\end{aligned}
\right.
\end{equation}
The temporal uncertainty term is:
\begin{equation}
u_{\text{time}}(v) = 
\beta_1 |\Delta S_t(v)| + 
\beta_2 \cdot 
\mathbb{I}(v \in M_{\text{focus}}),
\label{eq:temporal_uncertainty}
\end{equation}
where 
$M_{\text{focus}} = 
M_{\text{new}} \cup 
M_{\text{change}} \cup 
M_{\text{free}}$.

Then, the final hierarchical uncertainty is fused as:
\begin{equation}
u_{\text{final}}(v) = 
\alpha_1 u_{\text{imp}}(v) + 
\alpha_2 u_{\text{exp}}(v) + 
\alpha_3 u_{\text{time}}(v),
\label{eq:final_uncertainty}
\end{equation}
where $\alpha_i$ are weights estimated via evidence maximization, interpreted as a fusion of global priors, local observations, and temporal consistency. This \textbf{hierarchical map} directly links to the NBV reward in Eq.~\ref{eq:nbv_reward}, providing a multi-scale uncertainty signal that balances exploration coverage and model fidelity. 

\subsection{Next-Best-View Searching}
\label{sec:active_planning}
With the hybrid scene state $\mathcal{M}_k = \{\mathcal{F}_\theta, \mathcal{G}_k\}$ and hierarchical uncertainty map $u_{\text{final}}(v)$ defined in Eq.~\ref{eq:final_uncertainty}, the goal of active planning is to select the next viewpoint $\mathbf{c}_i$ that maximizes the expected \emph{hybrid information gain}.

\paragraph{EHIG Objective.}
Based on the final hierarchical uncertainty, we cast NBV selection as:
\begin{equation}
\mathbf{c}_i^* = 
\arg\max_{\mathbf{c} \in \mathcal{C}}
\mathbb{E} \left[ 
\Delta \mathcal{I}_{\text{hybrid}}(\mathbf{c}) 
\right],
\label{eq:nbv_eig}
\end{equation}
where $\mathcal{C}$ is the candidate viewpoint set, and $\Delta \mathcal{I}_{\text{hybrid}}$ measures the reduction of hybrid entropy:
\begin{equation}
\Delta \mathcal{I}_{\text{hybrid}}(\mathbf{c}) = 
\Delta H[\mathcal{F}_\theta] + 
\Delta H[\mathcal{G}_k],
\end{equation}
corresponding to global implicit and local explicit uncertainty reduction, respectively.

\paragraph{Voxel-wise Information Weighting.}
For a voxel $v$ visible from candidate $\mathbf{c}$, we define its contribution as:
\begin{equation}
w(v|\mathbf{c}) = 
\alpha U(v) + \beta H_{\text{hybrid}}(v),
\label{eq:nbv_weight}
\end{equation}
where $U(v)$ is the hierarchical uncertainty estimate from Eq.~\ref{eq:final_uncertainty} and $H_{\text{hybrid}}(v)$ is the hybrid entropy in Eq.~\ref{eq:hybrid_entropy}. $\alpha$ and $\beta$ are weights. This formulation unifies multi-scale uncertainty into a single information-theoretic weight.

\paragraph{NBV Reward.} Given the information weight of the voxel, the expected reward of candidate $\mathbf{c}$ is obtained via Eq.~\ref{eq:nbv_reward}.

\paragraph{Risk-Aware Path Planning.}
After obtaining the next goal, we employs an enhanced RRT* algorithm~\cite{lavalle2001rapidly} for active path planning. To generate physically feasible trajectories, we integrate the NBV reward into a risk-aware cost function:
\begin{equation}
\mathbf{p}^* = 
\arg\min_{\mathbf{p}}
\int_{\mathbf{p}}
\Big( C_{\text{travel}}(x) - 
\eta R(\mathbf{c}_x) + 
\lambda C_{\text{risk}}(x) \Big) dx,
\label{eq:risk_path}
\end{equation}
where $\mathbf{p}$ is the planned path, $C_{\text{travel}}$ the navigation cost, $C_{\text{risk}}$ the collision probability, and $R(\mathbf{c}_x)$ the NBV reward at pose $x$.

This proposed NBV searching bridges hybrid scene representation, multi-scale uncertainty, and active trajectory optimization into a single expected information gain framework. By combining global implicit entropy reduction and local explicit observation gain, the planner achieves coverage-aware and detail-preserving exploration.

\subsection{Uncertainty-driven Keyframe Selection}
Unlike conventional keyframe strategies that are tightly coupled with temporal ordering, we propose a \textbf{Uncertainty-driven} selection criterion that anchors high-information observations in the hybrid scene state $\mathcal{M}_k$. Rather than merely ensuring temporal coverage, the proposed keyframes act as a \emph{sparse attention mechanism}, focusing optimization on regions where the hybrid uncertainty is maximized.


\paragraph{Viewpoint-Based Keyframe Selection.}
By decoupling keyframe selection from temporal sampling and binding it to viewpoint-space information gain, our method avoids redundant observations and focuses optimization capacity on spatially complementary views, which is crucial for active reconstruction. For a newly acquired RGB-D frame $\mathbf{S}_c$ with camera pose $\mathbf{T}_{c,w}$, we compute its viewpoint divergence relative to the active keyframe set $\mathcal{S}_{\text{KF}}$:
\begin{equation}
\delta_c = 
\min_{\mathbf{S}_j \in \mathcal{S}_{\text{KF}}}
d_{\text{view}}\big(\mathbf{T}_{c,w}, 
\mathbf{T}_{j,w}\big),
\end{equation}
where $d_{\text{view}}$ measures the viewpoint baseline in SE(3) space, combining angular separation and projected frustum overlap. 

Aggressive active motion planning may cause an agent to overskip salient textural structures, we introduce a dual-uncertainty intersection criterion. Define the \textit{high-uncertainty intersection set} as:
\begin{equation}
\mathcal{V}_{\text{high}} = \left\{ v \in \mathcal{V}_u \mid u_{\text{exp}}(v) > \tau_h \land u_{\text{imp}}(v) > \tau_h \right\}.
\end{equation}
For frame $\mathbf{S}_c$, we compute its \textit{uncertainty coverage ratio} $\rho_c$ as the fraction of $\mathcal{V}_{\text{high}}$ visible in its frustum:
\[
\rho_c = |\mathcal{V}_{\text{high}} \cap \mathcal{V}_{\text{vis}}(\mathbf{S}_c)|  \big/  |\mathcal{V}_{\text{vis}}(\mathbf{S}_c)|,
\]
with $\mathcal{V}_{\text{vis}}$ being the visible voxel set. A frame is promoted to keyframe if:
\begin{equation}
(\delta_c > \tau_{\text{view}}) \land 
(\Delta \mathcal{I}_{\text{hybrid}}(\mathbf{S}_c) > \tau_{\text{info}}) \land 
(\rho_c > \tau_{\rho}),
\label{eq:kf_selection}
\end{equation}
where $\tau_{\text{view}}$, $\tau_{\text{info}}$, $\tau_{\rho}$ are viewpoint, information-gain and coverage threshold, respectively. The uncertainty-driven keyframe selection scheme actually establishes a sparse attention mechanism toward scene structures. This ensures selection of frames observing regions where both geometric and neural uncertainties are high.

\paragraph{Viewpoint-Space Sliding Window.}
Employing all keyframes for joint optimization still incurs excessive computational burden, prior approaches maintained a sliding window over continuous time. However, this strategy exhibits significant viewpoint redundancy as agent approaches the target, while failing to establish sufficient covisibility constraints upon revisiting similar locations. We maintain a local optimization window 
$\mathcal{W}_k = 
\{\mathbf{S}_{c_1}, \dots, \mathbf{S}_{c_m}\}$ 
indexed by spatially selected keyframes, not constrained by temporal adjacency. The hybrid state $\mathcal{M}_k$ is jointly refined via:
\begin{equation}
E_{\text{total}} = 
E_{\text{photo}} + 
E_{\text{geo}} + 
\lambda E_{\text{reg}},
\label{eq:sw_energy}
\end{equation}
where $E_{\text{photo}}$ enforces multi-view photometric consistency on $\mathcal{G}_k$, $E_{\text{geo}}$ aligns Gaussian primitives with the 
implicit SDF $\mathcal{F}_\theta$, and $E_{\text{reg}}$ prevents overfitting across non-overlapping viewpoints.

\section{Experiments}
\label{sec:experi}

\subsection{Implementation and Simulator}
We implement the proposed method within the Habitat simulator~\cite{savva2019habitat} as an active exploration system. The agent captures posed RGB-D observations along planned viewpoints. The camera field-of-view is set to $60^\circ$ vertically and $90^\circ$ horizontally, and the system processes sequences online with on-policy planning and incremental reconstruction. Further implementation details are presented in the supplementary material.

\subsection{Datasets, Metrics, and Baselines}
Following prior active mapping benchmarks~\cite{yan2023active}, 
we evaluate on two widely used datasets:(i) \textbf{Replica}~\cite{straub2019replica} with $8$ indoor scenes, and (ii) \textbf{Matterport3D (MP3D)}~\cite{chang2017matterport3d} with $5$ large-scale scenes exhibiting significant occlusion and spatial complexity.
All methods are run for $2000$ exploration steps on Replica and MP3D.

We report metrics targeting the critical objectives of active reconstruction: 
\emph{accuracy} (Acc, cm), \emph{completion} (Com, cm), and \emph{completion ratio} (C.R., \%), where Acc/Com are computed with a $5 cm$ threshold. 
To evaluate rendering quality, we report PSNR, SSIM, and LPIPS on held-out viewpoints. For additional geometric consistency analysis, we compute the Mean Absolute Distance (MAD) between the reconstructed SDF and ground-truth surfaces.

We compare our method against state-of-the-art active reconstruction frameworks:
ActiveNR~\cite{yan2023active}, ANM-S~\cite{kuang2024active}, 
NARUTO~\cite{feng2024naruto}, and ActiveSplat~\cite{11037548}. 
We further compare passive baselines in the supplemental material. All baselines are re-trained and evaluated locally for fair comparison.

\begin{figure*}[t]
    \centering
    \includegraphics[width=\linewidth]{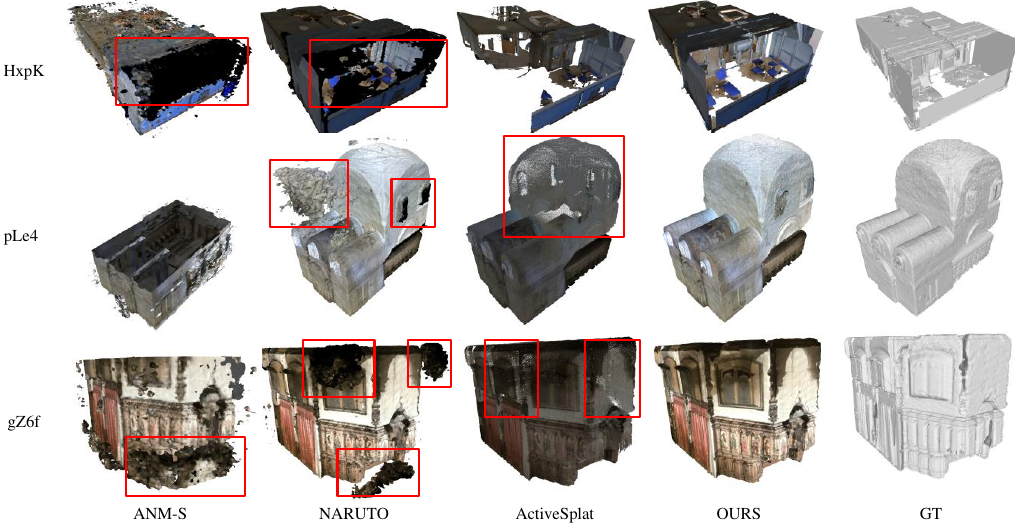}
    \caption{Qualitative comparison of 3D reconstruction results on representative MP3D sequences. 
    Additional results and detailed comparisons for all Replica and MP3D sequences are provided in the supplementary material.
    }
    \label{fig:mesh}
\end{figure*}

\begin{figure*}[t]
    \centering
    \includegraphics[width=\linewidth, trim={0 0 5 0},clip]{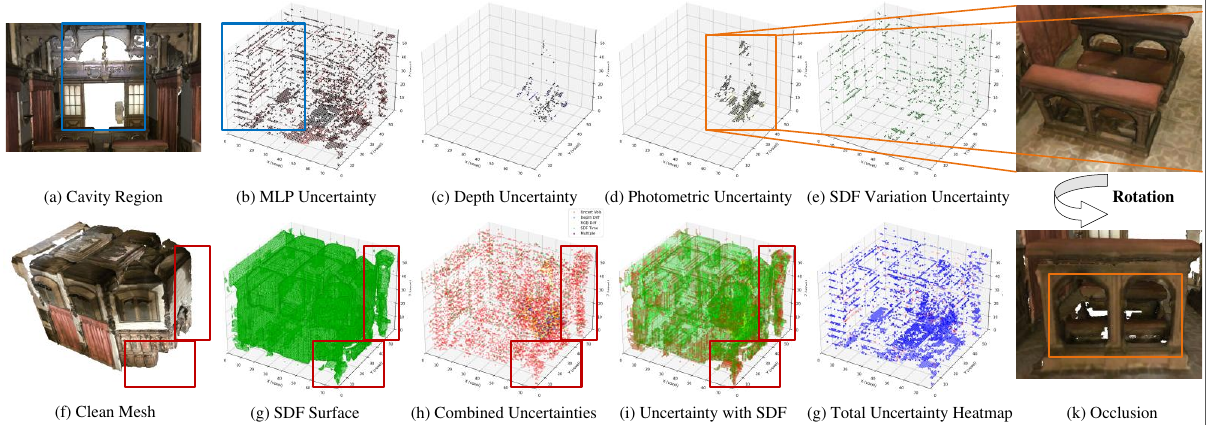}
    \caption{Visualization of uncertainties and their spatial relationship to real scene. Our proposed hybrid strategy not only endows the agent with global optimization capabilities, but also enables it to perceive intricate structures and textures while handling occlusions.}
    \label{fig:uncertainty}
\end{figure*}

\begin{table}[t]
\centering
 \resizebox{\linewidth}{!}{
\begin{tabular}{llcccccccc}
\toprule
Method & Metric & Off0 & Off1 & Off2 & Off3 & Off4 & R0 & R1 & R2 \\

\midrule
\multirow{3}{*}{ANM-S} 
& Acc (cm) $\downarrow$ & 1.44 & 1.03 & 1.60 & 1.80 & 1.50 & \cellcolor{blue!25}1.47 & 1.29 & 1.28    \\
& Com. (cm) $\downarrow$ & 1.98 & 1.55 & 6.65 & \cellcolor{blue!25}1.13 & 1.08 & 0.91 & 1.02 & \cellcolor{blue!25}0.85   \\
& C.R. (\%) $\uparrow$ & 95.43 & 92.66 & 79.20 & \cellcolor{blue!25}94.98 & 95.35 & 96.71 & 95.66 & 96.79   \\

\midrule
\multirow{6}{*}{NARUTO} 
& Acc (cm) $\downarrow$ & 1.26 & 1.04 & $\times$ & 34.84 & 1.67 & 1.75 & $\times$ & 1.50       \\
& Com. (cm) $\downarrow$ & 1.41 & 1.30 & $\times$ & 2.96 & 2.01 & 1.56 & $\times$ & 1.49       \\
& C.R. (\%) $\uparrow$ & 97.63 & 96.88 & $\times$ & 91.27 & 95.14 & 94.58 & $\times$ & 97.56       \\
& PSNR $\uparrow$ & 31.01 & 31.43 & $\times$ & 26.63 & 28.57 & 26.55 & $\times$ & 25.56       \\              
& SSIM $\uparrow$ & 0.892 & 0.897 & $\times$ & 0.831 & 0.882 & 0.782 & $\times$ & 0.818       \\  
& LPIPS $\downarrow$ & 0.299 & 0.283 & $\times$ & 0.283 & 0.284 & 0.354 & $\times$ & 0.367       \\

\midrule
\multirow{6}{*}{ActiveSplat} 
& Acc (cm) $\downarrow$ & 1.16 & 1.11 & 1.47 & 1.70 & 1.50 & 1.67 & 1.43 & 1.36    \\
& Com. (cm) $\downarrow$ & \cellcolor{blue!25}0.63 & \cellcolor{blue!25}0.94 & 5.59 & 1.83 & \cellcolor{blue!25}1.06 &\cellcolor{blue!25} 0.84 &\cellcolor{blue!25} 0.74 & 1.04    \\
& C.R. (\%) $\uparrow$ & 97.54 & 94.54 & 80.69 & 91.49 & 95.34 & 97.04 & 96.84 & 95.65    \\
& PSNR $\uparrow$ & 24.487 & 26.955 & 22.728 & 20.965 & 27.887 & 26.163 & 29.005 & 28.865    \\           
& SSIM $\uparrow$ & 0.857 & 0.871 & 0.888 & 0.804 & 0.878 & 0.823 & 0.877 & 0.894    \\
& LPIPS $\downarrow$ & 0.145 & 0.130 & 0.113 & 0.232 & 0.147 & 0.199 & 0.136 & 0.113    \\

\midrule
\multirow{6}{*}{OURS} 
& Acc (cm) $\downarrow$ & \cellcolor{blue!25}1.12 & \cellcolor{blue!25}1.02 & \cellcolor{blue!25}1.34 & \cellcolor{blue!25}1.56 & \cellcolor{blue!25}1.38 & 1.59 & \cellcolor{blue!25}1.13 & \cellcolor{blue!25}1.26   \\
& Com. (cm) $\downarrow$ & 1.34 & 1.17 & \cellcolor{blue!25}1.66 & 1.97 & 1.87 & 1.75 & 1.32 & 1.52    \\
& C.R. (\%) $\uparrow$ & \cellcolor{blue!25}97.76 & \cellcolor{blue!25}98.21 & \cellcolor{blue!25}96.86 & 94.70 & \cellcolor{blue!25}96.80 & \cellcolor{blue!25}97.28 & \cellcolor{blue!25}98.09 & \cellcolor{blue!25}98.18    \\
& PSNR $\uparrow$ & \cellcolor{blue!25}40.51 & \cellcolor{blue!25}40.54 & \cellcolor{blue!25}{33.72} & \cellcolor{blue!25}{34.14} & \cellcolor{blue!25}{37.37} & \cellcolor{blue!25}{33.80} & \cellcolor{blue!25}{34.63} & \cellcolor{blue!25}{36.00}       \\      
& SSIM $\uparrow$ & \cellcolor{blue!25}{0.980} & \cellcolor{blue!25}{0.979} & \cellcolor{blue!25}0.951 & \cellcolor{blue!25}0.949 & \cellcolor{blue!25}0.964 & \cellcolor{blue!25}0.948 & \cellcolor{blue!25}0.954 & \cellcolor{blue!25}0.962       \\  
& LPIPS $\downarrow$ & \cellcolor{blue!25}0.030 & \cellcolor{blue!25}0.034 & \cellcolor{blue!25}0.067 & \cellcolor{blue!25}0.075 & \cellcolor{blue!25}0.054 & \cellcolor{blue!25}0.072 & \cellcolor{blue!25}0.056 & \cellcolor{blue!25}0.053       \\  

\bottomrule                          
\end{tabular}}
\caption{Quantitative comparison of 3D reconstruction and view synthesis quality between the proposed method and state-of-the-art approaches on the Replica dataset. The symbol $\times$ indicates that the method fails to complete exploration within five trials.}
\label{table:sparse_rendering_recon}
\end{table}

\subsection{Evaluation on Replica}

Table~\ref{table:sparse_rendering_recon} reports 3D reconstruction and view synthesis metrics on the Replica dataset. Our method consistently achieves the best or second-best performance across all metrics. For reconstruction, it yields the highest completion ratio (C.R.) and lowest Acc/Com error, reaching 98.09\% C.R. on R1 and 98.18\% on R2. For view synthesis, it achieves the highest PSNR (up to 40.51) and SSIM (0.980) while maintaining the lowest LPIPS, demonstrating sharp textures and 
photometric consistency.

\subsection{Evaluation on MP3D}

\begin{table}[!t]
\centering
 \resizebox{\linewidth}{!}{
\begin{tabular}{llcccccc}
\toprule
Method & Metric & Gdvg & gZ6f & HxpK & pLe4 & YmJk & Avg. \\

\midrule

\multirow{3}{*}{ActiveINR} 
& Acc (cm) $\downarrow$ & 5.09 & 4.15 & 15.60 & 5.56 & 8.61  & 7.80 \\
& Com. (cm) $\downarrow$ & 5.69 & 7.43 & 15.96 & 8.03 & 8.46  & 9.11 \\
& C.R. (\%) $\uparrow$ & 80.99 & 80.68 & 48.34 & 76.41 & 79.35 & 73.15 \\

\midrule
\multirow{3}{*}{ANM-S} 
& Acc (cm) $\downarrow$ & 5.52 & 1.62 & 2.13 & 4.54 & 4.50 & 3.66 \\
& Com. (cm) $\downarrow$ & 3.95 & 2.01 & 12.49 & 2.51 & 3.53 & 4.90 \\
& C.R. (\%) $\uparrow$ & 91.00 & 94.58 & 60.39 & 95.02 & 88.65 & 85.93 \\

\midrule
\multirow{7}{*}{NARUTO}
& Acc (cm) $\downarrow$ & 2.34 & 3.57 & 7.29 & 4.46 & 9.52 & 5.44 \\
& Com. (cm) $\downarrow$ & 4.93 & 2.47 & 2.84 & 3.14 & 5.68 & 3.81 \\
& C.R. (\%) $\uparrow$ & 84.88 & 93.26 & 92.15 & 82.67 & 78.99 & 86.39 \\
& PSNR $\uparrow$ & 23.42 & 23.84 & 23.32 & 27.15 & 23.64 & 24.27 \\
& SSIM $\uparrow$ & 0.742 & 0.719 & 0.734 & 0.767 & 0.735 & 0.739 \\
& LPIPS $\downarrow$ & 0.416 & 0.523 & 0.492 & 0.554 & 0.517 & 0.500 \\

\midrule
\multirow{7}{*}{ActiveSplat} 
& Acc (cm) $\downarrow$ & 2.39 & \cellcolor{blue!25}1.74 & 2.53 & 4.09 & 9.52 & 4.05 \\
& Com. (cm) $\downarrow$ & 3.76 &\cellcolor{blue!25}1.34 & 24.28 & \cellcolor{blue!25}1.07 & 2.84 & 6.66 \\
& C.R. (\%) $\uparrow$ & 92.11 & 97.61 & 44.45 & \cellcolor{blue!25}99.10 & 90.78 & 84.81 \\
& PSNR $\uparrow$ & 22.77 & 16.40 & 18.33 & 23.49 & 24.57 & 21.12 \\
& SSIM $\uparrow$ & 0.700 & 0.601 & 0.776 & 0.667 & 0.852 & 0.719 \\
& LPIPS $\downarrow$ & 0.264 & 0.342 & 0.236 & 0.345 & \cellcolor{blue!25}0.156 & 0.269 \\

\midrule
\multirow{7}{*}{OURS} 
& Acc (cm) $\downarrow$ & \cellcolor{blue!25}1.68 & 1.90 & \cellcolor{blue!25}1.61 & \cellcolor{blue!25}2.68 & \cellcolor{blue!25}2.66 & \cellcolor{blue!25}2.11 \\
& Com. (cm) $\downarrow$ & \cellcolor{blue!25}1.59 & 1.96 & \cellcolor{blue!25}2.09 & 2.38 & \cellcolor{blue!25}2.81 & \cellcolor{blue!25}2.27 \\
& C.R. (\%) $\uparrow$ & \cellcolor{blue!25}98.23 & \cellcolor{blue!25}97.94 & \cellcolor{blue!25}98.12 & 94.55 & \cellcolor{blue!25}91.73 & \cellcolor{blue!25}96.11 \\
& PSNR $\uparrow$ & \cellcolor{blue!25}31.12 & \cellcolor{blue!25}32.43 & \cellcolor{blue!25}29.53 & \cellcolor{blue!25}33.14 & \cellcolor{blue!25}30.93 & \cellcolor{blue!25}31.43 \\
& SSIM $\uparrow$ & \cellcolor{blue!25}0.912 & \cellcolor{blue!25}0.939 & \cellcolor{blue!25}0.905 & \cellcolor{blue!25}0.920 & \cellcolor{blue!25}0.923 & \cellcolor{blue!25}0.920 \\
& LPIPS $\downarrow$ & \cellcolor{blue!25}0.160 & \cellcolor{blue!25}0.168 & \cellcolor{blue!25}0.176 & \cellcolor{blue!25}0.222 & 0.179 & \cellcolor{blue!25}0.181 \\

\bottomrule                          
\end{tabular}}

\caption{Quantitative comparison on the MP3D dataset for 3D reconstruction and novel view synthesis. }
\label{table:mp3d}
\end{table}

Table~\ref{table:mp3d} evaluates our method on the MP3D dataset. Compared to ActiveSplat, our approach significantly improves both geometry and rendering fidelity. We achieve the highest combined reconstruction score in nearly all scenes, exceeding 98\% on three out of five sequences. For photometric metrics, our method delivers the best PSNR and SSIM in four out of five cases, 
while maintaining the lowest LPIPS, reflecting perceptually consistent rendering.

Figure~\ref{fig:mesh} visualizes reconstructions on MP3D. Compared to NARUTO and ActiveSplat, our method produces sharper edges, fewer ghosting artifacts, and consistent textures under dynamic occlusion.

\subsection{Ablation Study}

As summarized in Table~\ref{table:ablation}, ablation studies are conducted on the challenging MP3D YmJk scene—characterized by significant occlusion and complex geometry. 
\paragraph{Uncertainty Setting.} Removing multi-resolution tri-plane encoding causes system failure due to complete loss of spatial perception. Eliminating the MLP-predicted uncertainty volume severely degrades reconstruction completeness (Com: 4.37 cm vs. 2.81 cm) by impeding global scene understanding. Exclusion of depth uncertainty induces erratic reconstruction (Acc: 4.75 cm vs. 2.66 cm) due to compromised surface fidelity estimation, which destabilizes optimization. Omission of RGB uncertainty substantially deteriorates rendering metrics (PSNR: 28.35 dB vs. 30.93 dB), attributable to degraded color/texture perception. Disabling the time-varying SDF representation markedly decreases reconstruction completeness. 

\paragraph{Searching and Planning.}Replacing the risk-aware path planner with naive uncertainty-volume aggregation degrades reconstruction coverage (C.R.: 89.23$\%$ vs. 91.73$\%$), as this suboptimal strategy prompts excessive surface proximity, reducing global observability while increasing collision risk. Finally, disabling keyframe management guided by spatial co-visibility and uncertainty underutilizes historical observations upon revisit, leading to rendering degradation.

\paragraph{Advantages of Hierarchical Uncertainties.}
Fig.~\ref{fig:uncertainty} visualizes the Hierarchical Uncertainty Map. The fully implicit uncertainty (b, e) provides the agent with global optimization capability. However, as the MLP-predicted SDF tends to generate redundant structures (f, g), it induces excessively high uncertainty in void regions (a) and redundant structure areas (g). This results in the agent allocating excessive attention to non-existent uncertainties (h). Conversely, the fully explicit uncertainty (c, d) aids the agent in identifying complex structures and textures. Nevertheless, due to its inability to perceive occluded regions (k) via $\alpha$-blending, it leads the agent to prematurely conclude optimization completeness and initiate subsequent planning. Our hybrid approach synergistically combines the strengths of both explicit and implicit representations. By adaptively weighting the explicit and implicit uncertainties, it enhances the agent's perceptual awareness across all local and global regions (g).

\begin{table}[t]
\centering
\resizebox{\linewidth}{!}{ 
\begin{tabular}{lccccccc}
\toprule
Method & PSNR$\uparrow$ & SSIM$\uparrow$ & LPIPS$\downarrow$ & MAD$\downarrow$ & Acc$\downarrow$ & Com$\downarrow$ & C.R.$\uparrow$ \\
\midrule
final & 30.93 & \cellcolor{blue!25}0.923 & \cellcolor{blue!25}0.179 & \cellcolor{blue!25}1.53 & 2.66 & \cellcolor{blue!25}2.81 & \cellcolor{blue!25}91.73 \\
w.o. Tri-plane Encoder & $\times$ & $\times$ & $\times$ & $\times$ & $\times$ & $\times$ & $\times$ \\
w.o. MLP Uncert & 30.89 & 0.917 & 0.191 & 1.80 & \cellcolor{blue!25}2.65 & 4.37 & 84.84 \\
w.o. Depth Uncert & 29.28 & 0.907 & 0.218 & 1.88 & 4.75 & 5.12 & 83.97 \\
w.o. RGB Uncert & 28.35 & 0.901 & 0.201 & 1.78 & 2.69 & 4.02 & 86.18 \\
w.o. SDF Temp & \cellcolor{blue!25}31.23 & 0.921 & 0.187 & 1.58 & 2.71 & 3.11 & 90.83 \\
w.o. Risk Planning & 30.78 & 0.916 & \cellcolor{blue!25}0.179 & 1.61 & 2.67 & 3.40 & 89.23 \\
w.o. Uncert Keyframe & 29.43 & 0.917 & 0.182 & 1.54 & 2.77 & 2.88 & 88.91 \\
w. Temporal Sliding Window & 28.69 & 0.910 & 0.186 & 1.62 & 2.69 & 3.72 & 87.02 \\
\bottomrule                          
\end{tabular}}
\caption{Ablation study on MP3D dataset. The best results are highlighted in the table.}
\label{table:ablation}
\end{table}

\section{Conclusion}
We have introduced Active3D, an active 3D reconstruction framework that unifies implicit neural fields and explicit Gaussian primitives into a hybrid information-theoretic formulation. 
By deriving a hierarchical uncertainty volume from this hybrid scene state, our method simultaneously captures global structural priors and local observation confidence, enabling principled next-best-view selection. 
An uncertainty-driven keyframe selection strategy anchors high-entropy viewpoints as sparse attention nodes, while a viewpoint-space sliding window performs uncertainty-aware local refinement to maintain global–local consistency. 
Formulating NBV planning as an Expected Hybrid Information Gain problem with a risk-aware path planner further ensures efficient and safe exploration. 

\begin{figure*}
    \centering
    \includegraphics[width=\linewidth]{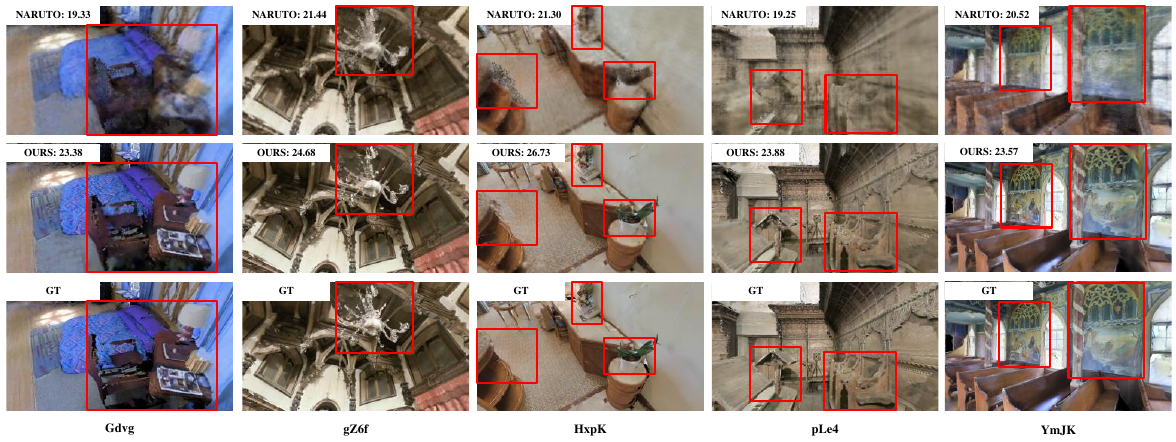}
    \caption{Novel view synthesis results on the MP3D dataset. The tested viewpoints were not present in any training trajectories of the evaluated methods. PSNR values are indicated in the top-left corner. Challenging regions are highlighted with red boxes.}
    \label{fig:render_mp3d}
\end{figure*}
\section*{Acknowledgments}
This research was supported by the Tier 2 Grant (MOE-T2EP20124-0015) from the Singapore Ministry of Education.

\input{CameraReady/LaTeX/appendix}

\bibliography{aaai2026}

\end{document}

%% file: CameraReady/LaTeX/appendix.tex
\appendix

\section{Supplementary Details}

This section elaborates on algorithmic details and experimental results omitted from the main text. We begin by introducing loss functions for both explicit and implicit reconstruction. Subsequently, we present computational efficiency metrics for each submodule and analyze the convergence point of reconstruction completeness during iterative refinement. Finally, extensive comparative evaluations against SOTA methods are provided, assessing reconstruction performance and rendering quality.

\subsection{Hybrid-map Optimization}

Within BA optimization, we compute gradients of the loss function with respect to both Gaussian parameters ($G_i = (\mu_i, \Sigma_i, \alpha_i, c_i)$) and the implicit weights of the MLP.

\begin{figure}[!t]
    \centering
    \includegraphics[width=\linewidth]{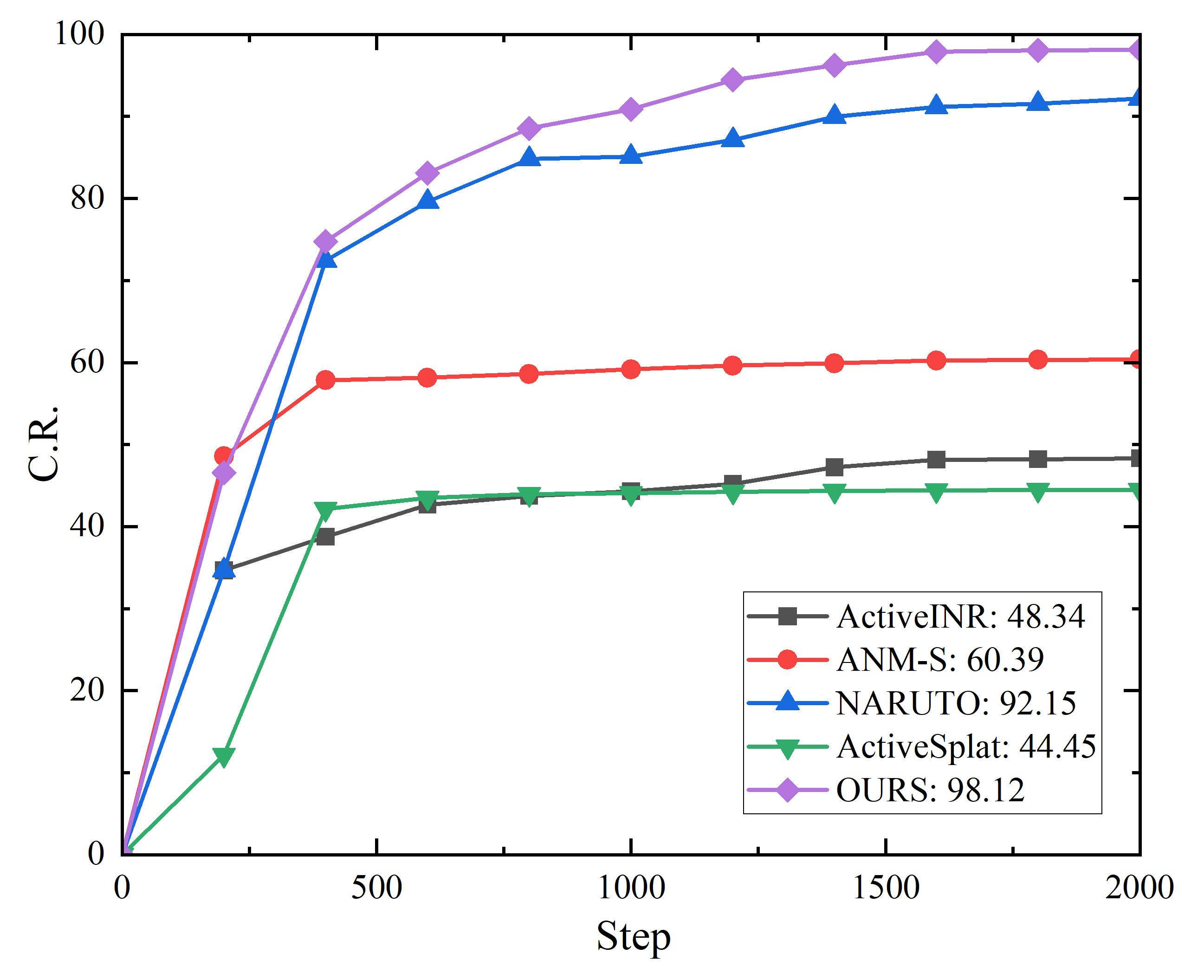}
    \caption{Convergence curves of completeness vs. iterations on the MP3D dataset. Our method achieves a faster convergence rate and higher final reconstruction completeness compared to other SOTA approaches.}
    \label{fig:coverge}
\end{figure}

\begin{table}[t]
\centering
\resizebox{\linewidth}{!}{ 
\begin{tabular}{lccccccc}
\toprule
Method & PSNR$\uparrow$ & SSIM$\uparrow$ & LPIPS$\downarrow$ & MAD$\downarrow$ & Acc$\downarrow$ & Com$\downarrow$ & C.R.$\uparrow$ \\
\midrule
Hybrid & \cellcolor{blue!25}32.43 & \cellcolor{blue!25}0.939 & \cellcolor{blue!25}0.168 & \cellcolor{blue!25}1.22 & \cellcolor{blue!25}1.90 & \cellcolor{blue!25}1.96 & \cellcolor{blue!25}97.94  \\
Implicit-only & 30.64 & 0.922 & 0.187 & 1.25 & 1.91 & 2.34 & 96.24\\
Explicit-only & 32.29 & 0.932 & 0.171 & 1.28 & 1.92 & 3.85 & 94.87 \\

\bottomrule                          
\end{tabular}}
\caption{Numerical Results of Hybrid Representation on MP3D dataset. The best results are highlighted in the table.}
\label{table:converge}
\end{table}

\begin{table*}
\centering
\resizebox{\linewidth}{!}{%
\begin{tabular}{lcccccccccccc}
\toprule
 & \multicolumn{3}{c}{Implicit Branch} & \multicolumn{2}{c}{Gaussian Splatting} & \multicolumn{3}{c}{Uncertainty Construction} & \multicolumn{4}{c}{Planning}\\
\cmidrule(lr){2-4} \cmidrule(lr){5-6} \cmidrule(lr){7-9} \cmidrule(lr){10-13} 
 & Ray Samping & MLP Inference & MLP Backward & Rendering & Mapping & Depth Uncert & Photometric Uncert & SDF Uncert & Uncert Aggre & NBV & Risk Field & RRT Planning \\
\midrule
Time (ms) & 6.56 & 7.38 & 14.46 & 7.03 & 35.42 & 6.34 & 7.55 & 2.55 & 16.34 & 1.98 & 2.12 & 3.07 \\
\bottomrule                          
\end{tabular}}
\caption{Computational Time Breakdown for System Components. The data represent the mean values derived from 2000 test frames in the MP3D dataset.}
\label{table:time}
\end{table*}

\begin{figure*}
    \centering
    \includegraphics[width=\linewidth]{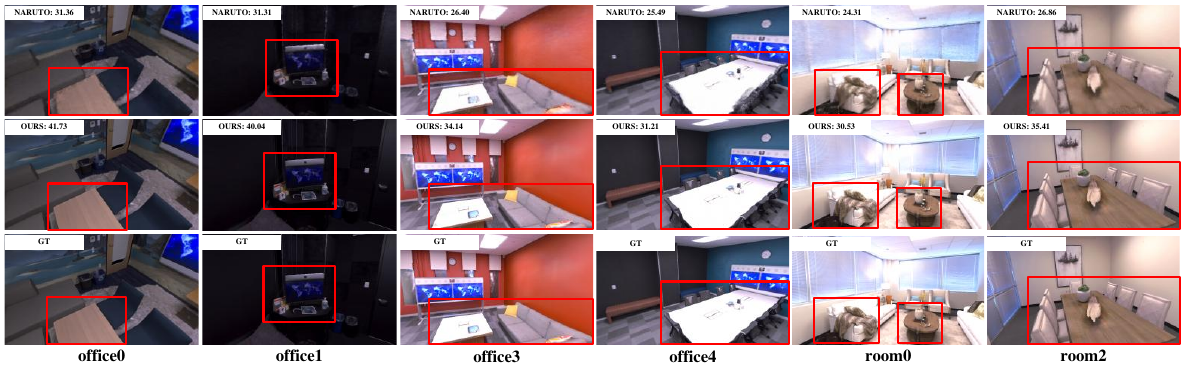}
    \caption{Novel view synthesis results on the Replica dataset. The tested viewpoints were not present in any training trajectories of the evaluated methods. PSNR values are indicated in the top-left corner. Challenging regions are highlighted with red boxes. The office2 and room1 sequences are not exhibited due to NARUTO's complete failure.}
    \label{fig:rende_replica}
\end{figure*}

\noindent\textbf{Explicit Loss.} 
In the explicit branch, each Gaussian primitive's parameters are optimized by minimizing photometric ($\mathcal{L}_{\text{pho}}$) and geometric ($\mathcal{L}_{\text{geo}}$) residuals between rendered and observed data:
\begin{equation}
    \begin{split}
        \mathcal{L}_{\text{pho}} & = \left\| I(\bm{\mathcal{M}}, \mathbf{T}_{c,w}, \mathbf{K}) - \bar{I} \right\|_2 \\
        \mathcal{L}_{\text{geo}} & = \left\| D(\bm{\mathcal{M}}, \mathbf{T}_{c,w}, \mathbf{K}) - \bar{D} \right\|_2
    \end{split}
    \label{eq:loss_pho_geo}
\end{equation}
where $\bar{I}$ and $\bar{D}$ represent the observed RGB image and depth map, while $I(\cdot)$ and $D(\cdot)$ denote the rendered images synthesized from the static Gaussian map $\bm{\mathcal{M}}$, camera pose $\mathbf{T}_{c,w} \in \mathrm{SE}(3)$, and intrinsic matrix $\mathbf{K} \in \mathbb{R}^{3\times3}$.

In the implicit branch, we employ four core loss functions to jointly optimize geometry, appearance, and uncertainty:

\noindent\textbf{RGB Loss.} 
For ray $i$ with rendered color $\hat{C}_i$ and ground truth $\bar{C}_i$, we compute:
\begin{equation}
\mathcal{L}_{\text{rgb}} = \frac{1}{N} \sum_{i=1}^{N} \omega_i \left\| \hat{C}_i - \bar{C}_i \right\|^2_2,
\end{equation}
where $N$ is the total ray count, and $\omega_i$ weights rays based on depth validity.

\noindent\textbf{Depth Loss.} 
For valid depth rays ($\bar{D}_j \leq D_{\text{trunc}}$):
\begin{equation}
\mathcal{L}_{\text{depth}} = \frac{1}{N_v} \sum_{j=1}^{N_v} \left( \hat{D}_j - \bar{D}_j \right)^2,
\end{equation}
where $N_v$ is the count of valid depth measurements, and $\hat{D}_j$ is the rendered depth for ray $j$.

\noindent\textbf{SDF Loss.} 
Given sampled depths $\mathbf{z} = \{z_i\}_{i=1}^{N_s}$ along a ray with $N_s$ samples, ground truth depth $d^*$, and predicted SDF values $\mathbf{s} = \{s_i\}_{i=1}^{N_s}$:
\begin{equation}
\mathcal{L}_{\text{sdf}} = \frac{1}{|\mathcal{M}|} \sum_{i \in \mathcal{M}} \left( s_i - (d^* - z_i) \right)^2,
\end{equation}
where $\mathcal{M} = \{ i : |z_i - d^*| < \tau \}$ defines the truncation region with threshold $\tau$.

\noindent\textbf{Uncertainty Loss.} 
For valid depth rays:
\begin{equation}
\mathcal{L}_{\text{uncert}} = \frac{1}{N_v} \sum_{j=1}^{N_v} \left( \frac{(\hat{D}_j - \bar{D}_j)^2}{2\sigma_j^2} + \frac{1}{2} \log \sigma_j^2 \right),
\end{equation}
where $\sigma_j^2$ is the variance from the uncertainty grid for ray $j$. The Uncertainty Loss incorporates two regularization terms. The first term reflects depth prediction uncertainty, promoting uncertainty amplification when depth estimates deviate significantly from ground truth to enhance agent attentiveness, while attenuating uncertainty under accurate predictions. The second term serves to curb excessive uncertainty expansion.

\noindent\textbf{Total Loss.}
The unified objective combines all loss components:
\begin{equation}
\mathcal{L}_{\text{total}} = \sum_{k \in \{\text{pho}, \text{geo}, \text{rgb}, \text{depth}, \text{sdf}, \text{uncert}\}} \lambda_k \mathcal{L}_k,
\end{equation}
where $\lambda_k$ are balancing weights for each loss term.

\subsection{Computational Efficiency and Convergence}

Owing to our dual-branch framework comprising both implicit and explicit reconstructions, we demonstrate the real-time capability by separately measuring the forward inference and backward optimization latency per RGB-D frame, further analyzing the computational cost of four key uncertainty components, and presenting the efficiency of path planning.

\begin{figure*}
    \centering
    \includegraphics[width=\linewidth]{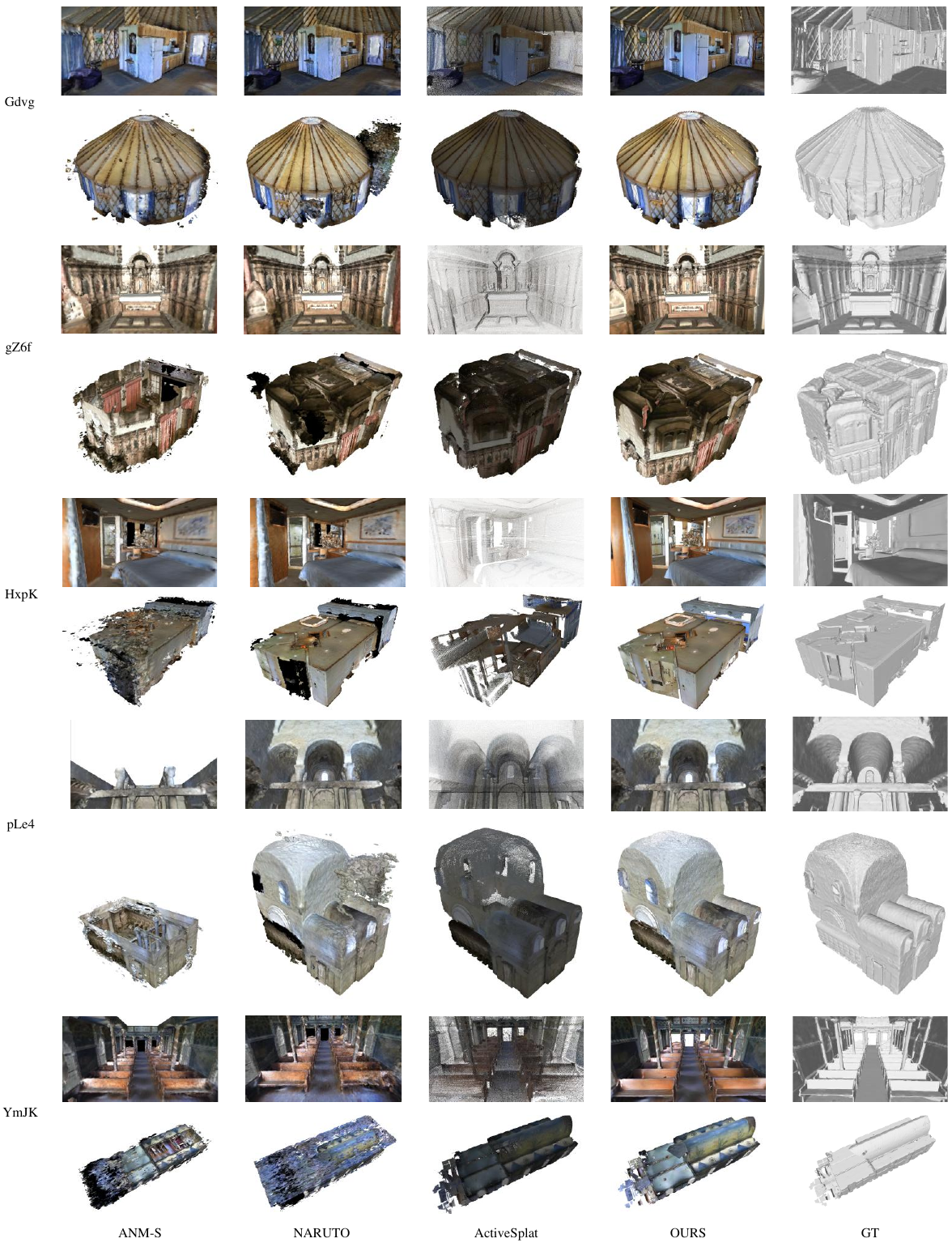}
    \caption{Reconstruction results on all 5 sequences of the MP3D dataset. The first row of each group illustrates local details, while the second row demonstrates global completeness. Our method achieves reconstructions with higher-fidelity local geometric details and superior completeness, while maintaining robustness across diverse scenarios. Scene appearance may exhibit variations due to method-specific simulator lighting configurations.}
    \label{fig:mesh_mp3d}
\end{figure*}

\begin{figure*}
    \centering
    \includegraphics[width=0.9\linewidth]{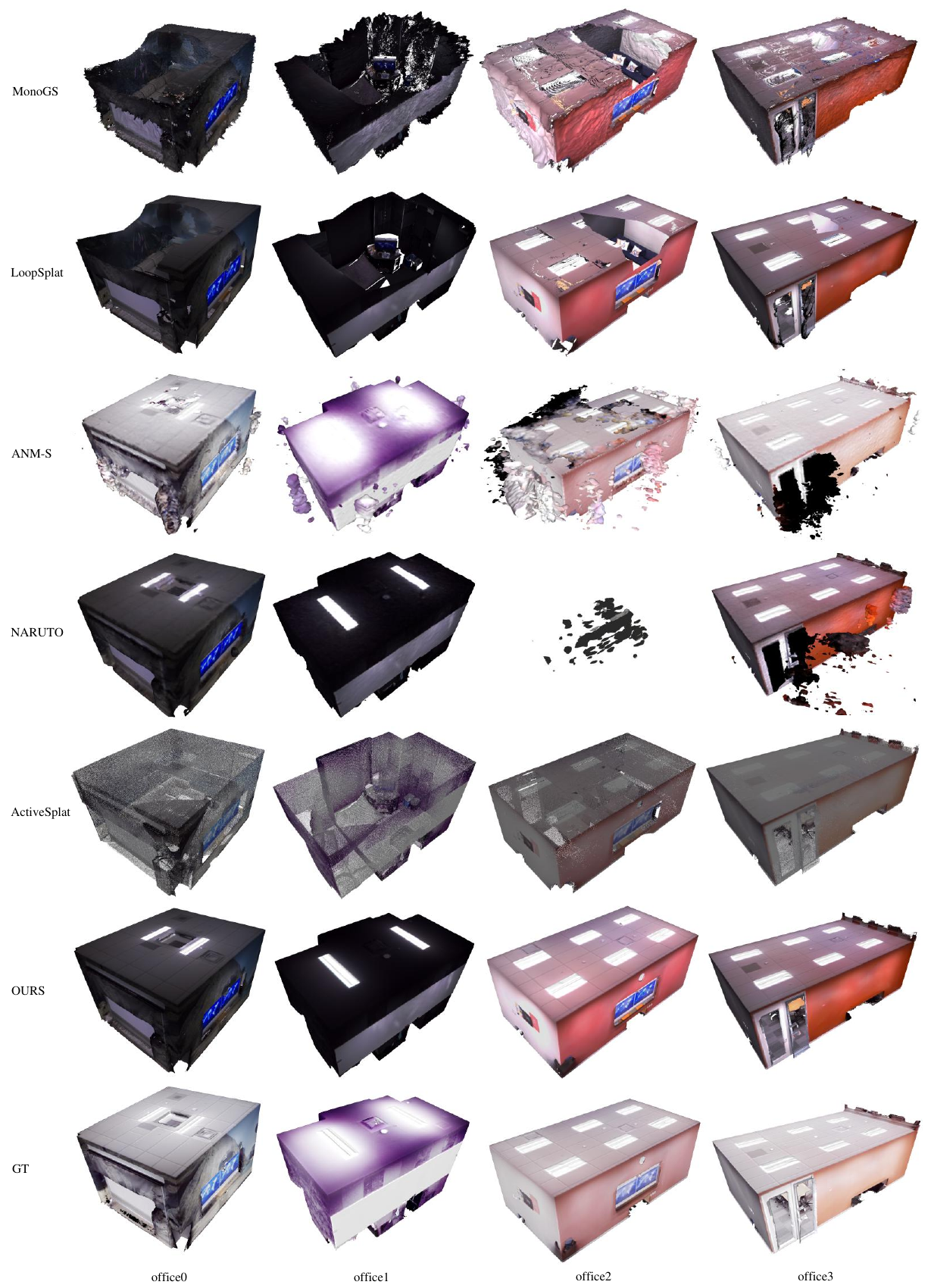}
    \caption{Reconstruction results on the first 4 sequences of the Replica dataset. While MonoGS and LoopSplat are passive reconstruction approaches, the others represent active reconstruction schemes. Our method reconstructs more complete scene structures and demonstrates robustness across all sequences. Scene appearance may exhibit variations due to method-specific simulator lighting configurations.}
    \label{fig:mesh_replica_1}
\end{figure*}

\begin{figure*}
    \centering
    \includegraphics[width=0.9\linewidth]{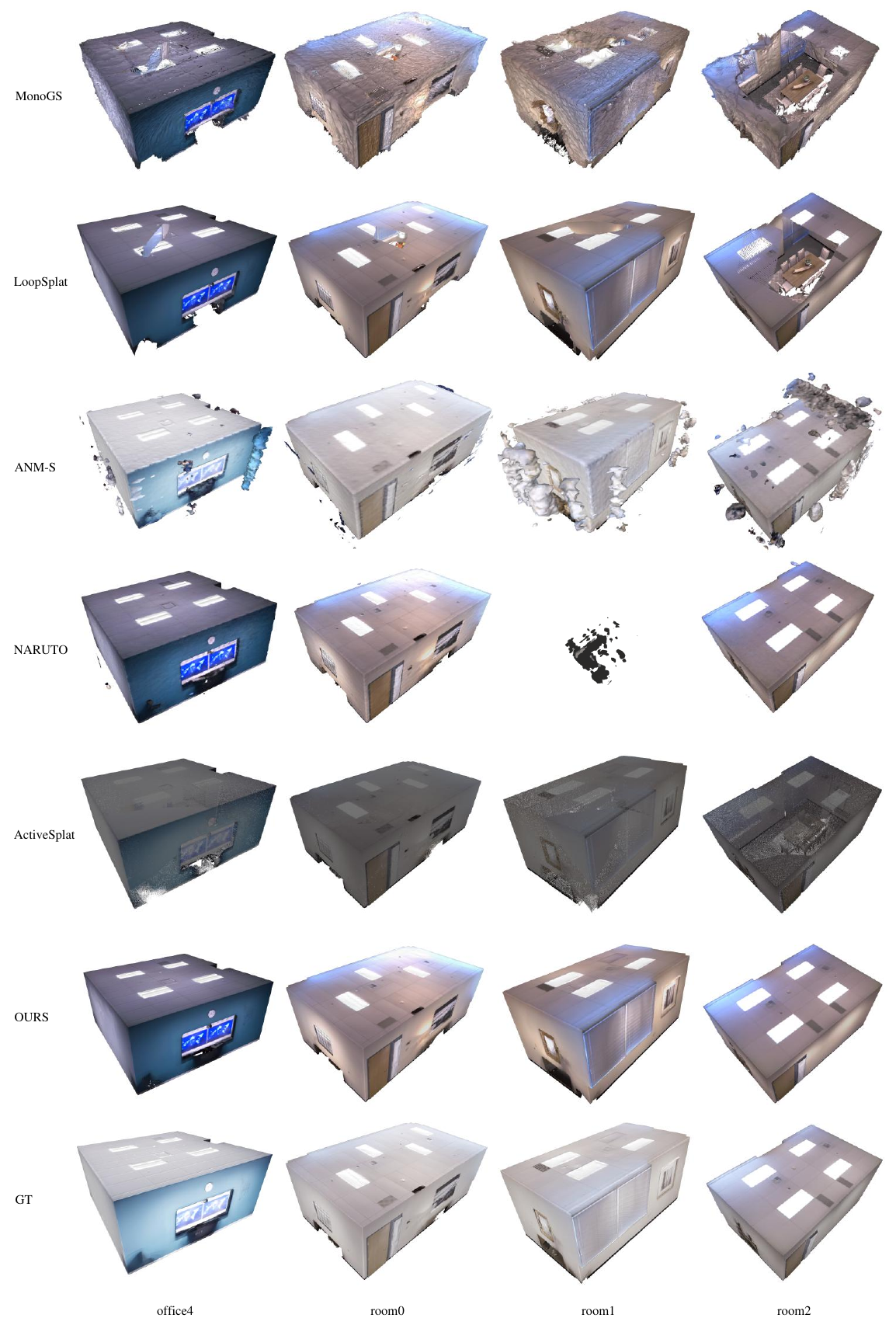}
    \caption{Reconstruction results on the last 4 sequences of the Replica dataset. While MonoGS and LoopSplat are passive reconstruction approaches, the others represent active reconstruction schemes. Our method reconstructs more complete scene structures and demonstrates robustness across all sequences. Scene appearance may exhibit variations due to method-specific simulator lighting configurations.}
    \label{fig:mesh_replica_2}
\end{figure*}

\noindent\textbf{Time Analysis.} 
Table~\ref{table:time} presents the average computation time for key modules. For the implicit branch, the primary computational cost lies in coordinate point sampling, model forward inference, and back-propagation. Our MLP is lightweight due to its shallow depth and minimal number of hidden neurons. Within the Gaussian reconstruction branch, significant time is consumed by $\alpha$-blending and Gaussian map optimization, which is performed only on keyframes within the sliding window to enhance computational efficiency. For the uncertainty voxels, implicit uncertainty voxels are directly obtained via MLP inference, while the construction of other uncertainty voxels completes within milliseconds. Finally, the path planning module aggregates uncertainty voxels, identifies the NBV, constructs a risk field, and performs RRT planning. Note that planning is triggered only when the agent reaches its target position and initiates the next planning cycle; the system primarily operates in motion execution, thereby enhancing overall efficiency.

\noindent\textbf{C.R. Convergence.} 
Fig.~\ref{fig:coverge} illustrates the reconstruction completeness curve versus iteration count. In challenging scenario \emph{YmJk}, ActiveSplat~\cite{11037548} fails to plan effective trajectories, significantly hindering its performance. While Active-INR~\cite{yan2023active} and ANM-S~\cite{kuang2024active} exhibit rapid initial exploration, they ultimately converge to low completeness levels. NARUTO~\cite{feng2024naruto} demonstrates superior convergence and completeness, yet it frequently predicts large-scale redundant structures in invalid regions (refer to reconstruction results in the main text). In contrast, our method robustly plans trajectories and reconstructs high-fidelity meshes.

\subsection{Rendering Performance}

Compared to NeRF, Gaussian Splatting exhibits superior novel view synthesis. Quantitative evaluations in the main text confirm our approach outperforms SOTA methods across all metrics. This subsection provides qualitative comparisons on all MP3D and Replica sequences. Fig.~\ref{fig:render_mp3d} and Fig.~\ref{fig:rende_replica} show results for MP3D and Replica, respectively. Notably, test viewpoints were excluded from all training trajectories. Our method renders sharp boundaries and clear textures.

\subsection{Reconstruction Results}

Mesh reconstructions for three MP3D sequences are compared with SOTA in the main text. Fig.~\ref{fig:mesh_mp3d}-\ref{fig:mesh_replica_2} show full qualitative comparisons across five MP3D and eight Replica sequences. MonoGS~\cite{matsuki2023gaussian} and LoopSplat~\cite{zhu2024loopsplat} (passive schemes) exhibit extensive fragmentation due to incomplete scene observation. Neural methods ANM-S~\cite{kuang2024active} and NARUTO~\cite{feng2024naruto} partially fill holes but generate superfluous structures, causing agent over-focus as shown in the text. ActiveSplat~\cite{11037548} produces only sparse point clouds. In contrast, our approach ensures superior global integrity and enhanced local geometric accuracy.

\section{Hybrid Representation Analysis}

To validate the hybrid implicit-explicit formulation, we compare against two variants: (i) \emph{implicit-only}, using only $\mathcal{F}_\theta$, and 
(ii) \emph{explicit-only}, using only $\mathcal{G}_k$. 

\subsection{Implicit-only}

For the implicit-only evaluation, we utilize the MLP-predicted SDF and uncertainty for active exploration. We simultaneously capture RGB-D images along the planned trajectory, perform Gaussian projection, and optimize the Gaussian branch using the loss defined in Eq.~\ref{eq:loss_pho_geo}. However, the explicit and implicit branches remain entirely independent, yielding no mutual enhancement or interference.
\subsection{Explicit-only}

For the explicit-only evaluation, we employ both Depth Uncertainty and Photometric Uncertainty to construct an uncertainty voxel map, computing the discrepancy between the ground-truth depth and the observed depth as the SDF map, which is updated in real-time during subsequent tracking. During incremental updates, voxel grids are projected into camera coordinates and validated against depth bounds and image constraints. Valid voxels sample depth values via bilinear interpolation, enforcing local depth continuity to reject outliers. The SDF observations are truncated to a narrow band around surfaces. Each voxel update includes depth adaptation and consistency weight between images, which is maintained by an exponential weighted moving average.

Although uncertainty predicted by the implicit MLP is not utilized, we retain its presence while optimizing the MLP for convergence analysis.

\subsection{Complementarity Analysis}

\begin{figure*}[t]
    \centering
    \includegraphics[width=\linewidth]{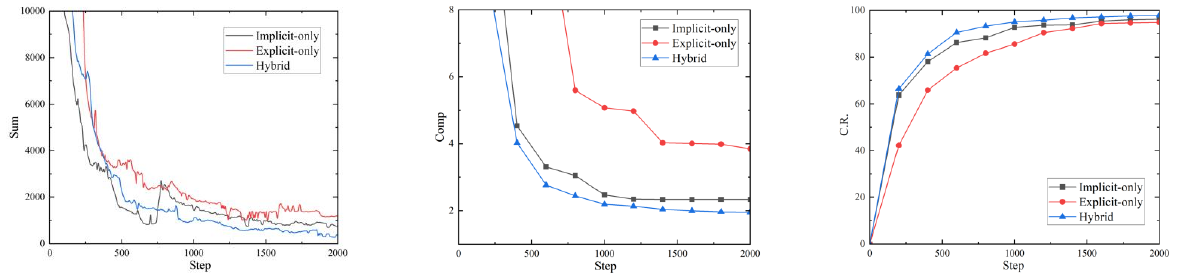}
    \caption{Convergence-divergence analysis of hybrid representation on MP3D dataset. The left figure depicts the evolution of predictive uncertainty from the MLP during the iterative process. The center figure plots the reconstruction accuracy against iteration count. The right figure illustrates the progression of reconstruction completeness throughout the iterations.}
    \label{fig:converge_uncert}
\end{figure*}

\noindent\textbf{Completeness Analysis.}
Fusing implicit and explicit branches yields a statistically significant acceleration in reconstruction convergence speed and elevates final reconstruction completeness. This synergistic integration facilitates mutual enhancement between the two map representation paradigms. Table~\ref{table:converge} shows the quantitative comparison of different representation modes.

\subsection{Risking Filed and Planning}

Conventional path planners rely on explicit representations (e.g., occupancy grids, octrees), yet suffer from degraded navigation accuracy in complex environments due to erroneous occupancy estimation. SDF fields conversely exhibit initial exploration instability, frequently guiding paths into obstacles and causing collisions. To address these limitations, we propose a hybrid implicit-explicit planner. Our framework constructs an occupancy field via Gaussian processes, integrates it with an SDF to generate a risk field, enabling efficient global exploration and planning.

\begin{figure}[t]
    \centering
    \includegraphics[width=\linewidth]{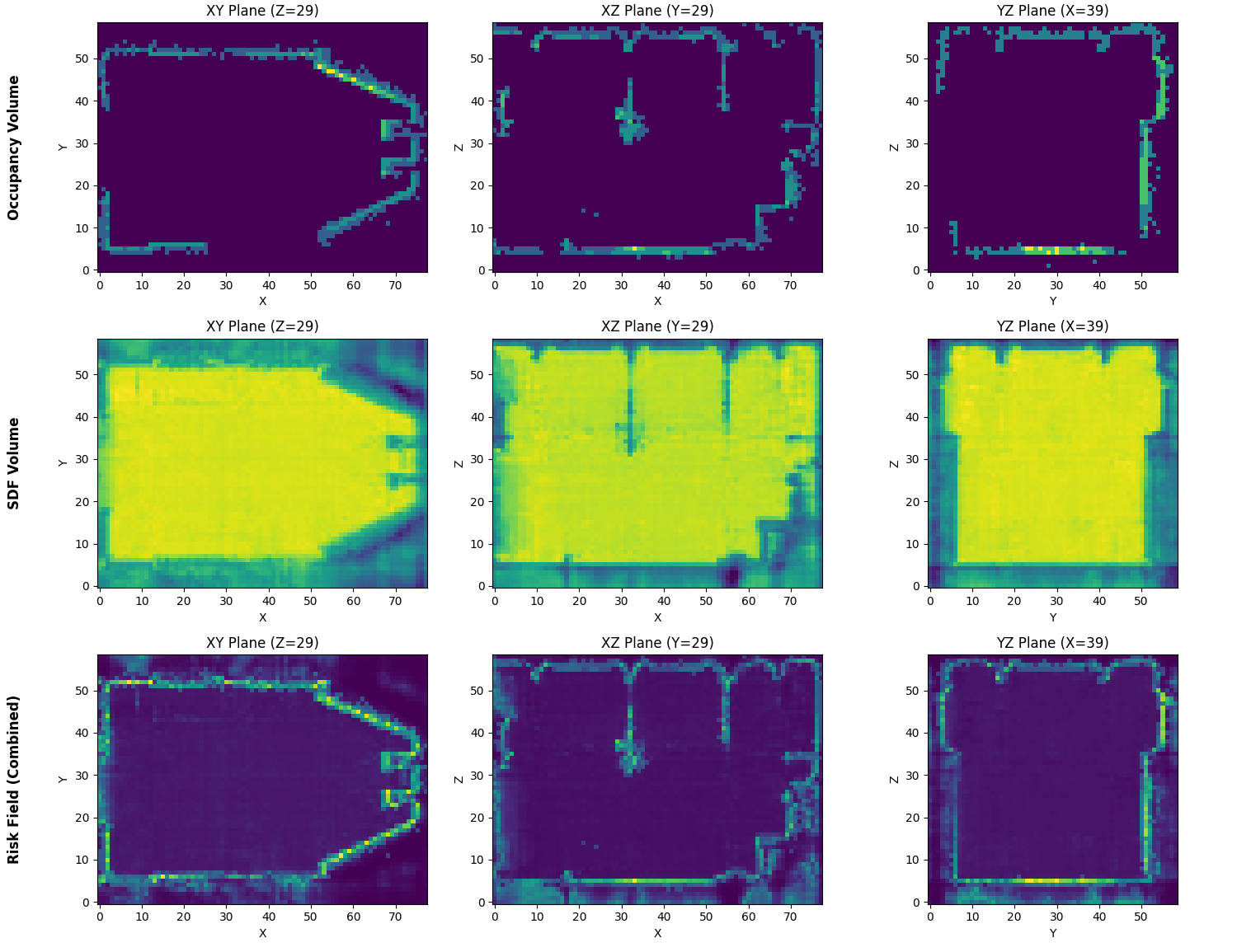}
    \caption{Visualization of occupancy field, sdf field, and risk field for MP3D gZ6f. The risk field completes the gaps within the occupancy field and eliminates the inherent redundant structures in the SDF field.}
    \label{fig:risk_filed}
\end{figure}

Fig.~\ref{fig:risk_filed} visualizes cross-sections of the occupancy field, SDF field, and risk field on MP3D sequence gZ6f. Initial exploration shows the Gaussian-based occupancy map is incomplete, misleading planners into boundary violations. The SDF map generates redundant structures. Our risk field synergistically preserves distinct object boundaries while eliminating redundant structures.

\noindent\textbf{Uncertainty Convergence Analysis.}
For the three schemes (Hybrid, Implicit-only, Explicit-only), we statistically analyzed the temporal evolution of the MLP-predicted uncertainty throughout the tracking and reconstruction pipeline. Fig.~\ref{fig:converge_uncert} juxtaposes these trends against the convergence/divergence of reconstruction accuracy (Comp) and completeness rate (C.R.). The uncertainty convergence curves reveal that the Implicit-only scheme culminates in persistently elevated uncertainty values. This stems from its tendency to over-prioritize redundant structures (as analyzed in the main text), thereby diminishing exploratory coverage in structurally rich regions and degrading reconstruction quality and completeness. Furthermore, gradient optimization of the uncertainty MLP exhibits direct correlation with depth prediction fidelity. Suboptimal reconstruction accuracy consequently amplifies uncertainty, establishing a detrimental bidirectional feedback loop.

\noindent\textbf{Reconstruction Accuracy Analysis.}
The Explicit-only scheme demonstrates consistently inferior accuracy across all iterations. This limitation arises from its fundamental inability to address occlusions, causing premature abandonment of under-optimized regions. In contrast, the integration of implicit uncertainty imbues the agent with heightened attentional focus on occluded areas, thereby enhancing reconstruction precision.